\documentclass{tlp}

\usepackage{amsmath}
\usepackage{graphicx}
\usepackage{multirow}
\usepackage{algorithm}
\usepackage{algorithmic}

\usepackage{amsmath}
\usepackage{amssymb} 

\usepackage{booktabs}
\usepackage{multirow}
\newtheorem{definition}{Definition}
\newtheorem{theorem}{Theorem}

\begin{document}

\lefttitle{Causality Constrained Counterfactual Explanations}

\jnlPage{TO DO}{TO DO}
\jnlDoiYr{2024}
\doival{TO DO}

\title[Causality Constrained Counterfactual Explanations]{CFGs: Causality Constrained Counterfactual Explanations using goal-directed ASP\thanks{This work was supported by the US NSF Grants IIS 1910131 and IIP 1916206, US DoD, grants from industry. We also thank the members of ALPS lab at UT Dallas for the insightful discussions.}}

\begin{authgrp}
\author{\sn{Sopam Dasgupta}}
\affiliation{The University of Texas at Dallas}
\author{\sn{Joaqu\'in Arias}}
\affiliation{CETINIA, Universidad Rey Juan Carlos}
\author{\sn{Elmer Salazar}}
\affiliation{The University of Texas at Dallas}
\author{\sn{Gopal Gupta}}
\affiliation{The University of Texas at Dallas}
\end{authgrp}

\history{\sub{13 05 2024;} \rev{13 05 2024;}}

\maketitle

\begin{abstract}
\vspace{-0.05 in}
Machine learning models that automate decision-making are increasingly used in consequential areas such as loan approvals, pretrial bail approval, and hiring. Unfortunately, most of these models are black boxes, i.e., they are unable to reveal how they reach these prediction decisions. A need for transparency demands justification for such predictions. An affected individual might also desire explanations to understand why a decision was made. Ethical and legal considerations require informing the individual of changes in the input attribute (s) that could be made to produce a desirable outcome. Our work focuses on the latter problem of generating \textit{counterfactual explanations} by considering the causal dependencies between features.
In this paper, we present the framework CFGs, \textit{CounterFactual Generation with s(CASP)}, which utilizes the goal-directed Answer Set Programming (ASP) system s(CASP) to automatically generate counterfactual explanations from models generated by \textit{rule-based machine learning} algorithms in particular. We benchmark CFGs with the FOLD-SE model. Reaching the counterfactual state from the initial state is planned and achieved using a series of interventions. To validate our proposal, we show how counterfactual explanations are computed and justified by imagining worlds where some or all factual assumptions are altered/changed. More importantly, we show how CFGs navigates between these worlds, namely, go from our initial state where we obtain an undesired outcome to the imagined goal state where we obtain the desired decision, taking into account the causal relationships among features.
\end{abstract}
\vspace{-0.1 in}

\begin{keywords}
Causal reasoning, Counterfactual reasoning, Default Logic, Answer Set Programming, Planning problem
\vspace{-0.1 in}
\end{keywords}
\vspace{-0.2 in}

\section{Introduction}
Predictive models are used in automated decision-making, for example, in the filtering process for hiring candidates for a job or approving a loan. Unfortunately, most of these models are like a black box, making understanding the reasoning behind a decision difficult. In addition, many of the decisions such models make have consequences for humans affected by them. Humans desire a satisfactory explanation when subject to unfavorable decisions, judgments, or outcomes. This desire for transparency is essential, regardless of whether an automated system (e.g., a data-driven prediction model) or other humans make such consequential decisions. Hence, making such decisions explainable to people is a challenge. For a decision made by a machine learning system, Wachter et al.\cite{wachter} highlight an approach where a counterfactual is generated to explain the reasoning behind a decision and inform a user on how to achieve a positive outcome. 

Our contribution in this paper is a framework called \textit{Counterfactual Generation with s(CASP) (CFGs)} that generates counterfactual explanations from \textit{rule-based machine learning (RBML)} algorithms. In doing so, we attempt to answer the question, ‘What can be done to achieve the desired outcome given that the outcome currently received is undesired?’. Our approach \textit{CFGs} models various worlds/scenarios---one is the current scenario/initial state where the machine learning model produces an undesired outcome and the other is the imagined scenario/goal state where the same model generates the desired outcome. Our approach guides us to reach the goal state from the initial state. 
The traversal between these worlds/states, i.e., from the original world/initial state where we got the undesired negative outcome to the counterfactual world(s)/goal state(s) where we get the desired positive decision, is done through \textit{interventions/transitions}, i.e., by changing \textit{feature values}. These interventions are made while taking causal dependencies between features into account. \textit{CFGs} relies on commonsense reasoning, as realized via answer set programming (ASP) (Gelfond \& Kahl\cite{gelfond-kahl}), specifically the goal-directed s(CASP) ASP system by Arias et al.\cite{scasp-iclp2018}. Specifically, it views the problem of finding the interventions as a planning problem (Gelfond \& Kahl\cite{gelfond-kahl}).


\vspace{-0.2in}

\section{Background}
\vspace{-0.03 in}
\subsection{Counterfactual Reasoning}
\vspace{-0.03 in}
As humans, we treat explanations as tools to help us understand decisions and inform us on how to act. Wachter et al.\cite{wachter} argued that counterfactual explanations (CFE) should be used to explain individual decisions. Counterfactual explanations offer meaningful explanations to understand a decision and inform on what can be done to change the outcome to a desired one. In the example of being denied a loan, counterfactual explanations are similar to the statement: If John had \textit{good} credit, his loan application would be approved. A key idea behind counterfactual explanations is imagining a different world/ state where the desired outcome would hold as well as be reachable from the current world. Thus, counterfactuals involve imagining alternate (reasonably plausible) worlds/scenarios where such a desired outcome would be achievable.

For a binary classifier used for prediction, given by $f:X \rightarrow \{0,1\}$, we define a set of counterfactual explanations $\hat{x}$ for a factual input $x \in X$ as $\textit{CF}_{f}(x)=\{\hat{x} \in X | f(x) \neq f(\hat{x})\}$. The set of counterfactual explanations contains all the inputs ($\hat{x}$) that lead to a prediction under  $f$ that differs from the original input $x$ prediction. 

Additionally, Ustun et al.\cite{ref_2_ustun} highlighted the importance of \textit{algorithmic recourse} for counterfactual reasoning, which we define as \textit{altering undesired outcomes by algorithms to obtain desired outcomes through specific actions}. Building on top of the work by Ustun et al.\cite{ref_2_ustun}, Karimi et al.\cite{alt_karimi} showed how to generate counterfactual explanations that consider the immutability of features. For example, a counterfactual instance that recommends changing the `gender' or `age' of an individual has limited utility. Their work allows restriction on the kind of changes that can be made to feature values in the process of generating plausible counterfactual explanations. Additionally, the work of both Ustun et al.\cite{ref_2_ustun} and Karimi et al.\cite{alt_karimi} generates a set of diverse counterfactual explanations. However, they assume that features are independent of each other. In the real world, there may be causal dependencies between features. 

We show how counterfactual reasoning is performed using the s(CASP) query-driven predicate ASP system (Arias et al.\cite{scasp-iclp2018}) while considering \textit{causal dependency} between features. Counterfactual explanations are generated by utilizing s(CASP)’s inbuilt ability to compute the \textit{dual rules} (as described in Section \ref{dual_rules}) that allow us to execute negated queries. Given the definition of a predicate {\tt p} as a rule in ASP, its corresponding \textit{dual rule} allows us to prove {\tt not p}, where {\tt not} represents \textit{negation as failure} (Lloyd\cite{ref_NAF_1}). We utilize these \textit{dual rules} to construct alternate worlds/states that lead to counterfactual explanations while considering causal dependencies between features. 

\vspace{-0.1 in}
\subsection{Causality}
\vspace{-0.05 in}

Inspired by the Structural Causal model approach \cite{SCM}, Karimi et al.\cite{ref_4_karimi_2} showed how only considering the nearest counterfactual explanations in terms of cost/distance, without taking into account causal relations that model the world, produce unrealistic explanations that are not realizable. Their work focused on generating counterfactual explanations through a series of readily achievable interventions that provide a realistic path to flipping the predicted label. In earlier approaches to generating counterfactuals by Ustun et al.\cite{ref_2_ustun} and Karimi et al.\cite{alt_karimi}, implicit assumptions are made where changes resulting from  interventions will be independent of changes across features. However, this is only true in worlds where features are independent. This assumption of independence across features might need to be revised in the real world. 

Causal Relationships that govern the world should be taken into account. For example, an individual wants to be approved for a loan. Given that the loan approval system takes into account the marital status, the counterfactual generation system, which takes into account the marital status, the relationship status, the gender, and age, might make a recommendation of changing the marital status value to `single.' However, the individual still has a relationship status of `husband.' A causal dependency exists between at least two features (marital status and relationship). In addition, the `gender' of the individual influences if the person is a `husband' or `wife.' Hence, a causal dependency exists between `gender' and `relationship'. Unless these causal dependencies are taken into account, a counterfactual explanation of changing one's marital status by assuming that relationship and gender are independent is unrealistic. It might not even result in the loan being approved (assuming the loan approval system considers relationship and marital status). Realistic counterfactual explanations can be generated by highlighting the importance of modeling causal relationships, which, in turn, model down-steam changes caused by directly changing features. 

\vspace{-0.15 in}

\subsection{ASP, s(CASP) and Commonsense Reasoning}\label{dual_rules}
\vspace{-0.05 in}

\textbf{Answer Set Programming (ASP)} is a well-established paradigm for knowledge representation and reasoning (Brewka et al.\cite{cacm-asp}; Baral\cite{baral}; Gelfond \& Kahl\cite{gelfond-kahl}) with many prominent applications, especially to automating commonsense reasoning. We use ASP to represent the knowledge of the features---the domains, the properties of the features, the decision-making rules, and the causal rules. We then utilize this encoded symbolic knowledge to generate counterfactual explanations automatically.

\textbf{s(CASP)} is a goal-directed ASP system that executes answer set programs in a top-down manner without grounding them (Arias et al.\cite{scasp-iclp2018}). The query-driven nature of s(CASP) greatly facilitates performing commonsense reasoning as well as counterfactual reasoning based on commonsense knowledge employed by humans. Additionally, s(CASP) justifies counterfactual explanations by utilizing proof trees. 
 
In s(CASP), to ensure that facts are true only as a result of following rules and not by any spurious correlations, \textit{program completion} was adopted, which replaces a set of ``if" rules with ``if and only if" rules. One way to implement this was for every rule that says $p \implies q$; we add a complementary rule saying $\neg p \implies \neg q$. The effect ensures that q is true ``if and only if" p is true.
In s(CASP), \textit{program completion} is done by introducing \textbf{dual rules}. For every rule of the form $p \implies q$, s(CASP) automatically generates dual rules of the form $\neg p \implies \neg q$ to ensure \textit{program completion}. 

\textbf{Commonsense reasoning} in ASP can be emulated using default rules, integrity constraints, and multiple possible worlds~(Gupta et al.\cite{ref_GG}; Gelfond \& Kahl\cite{gelfond-kahl}).  We assume the reader is familiar with ASP and s(CASP). An introduction to ASP can be found in Gelfond and Kahl's book \cite{gelfond-kahl}, while a detailed overview of the s(CASP) system can be found elsewhere (Arias et al.\cite{arias-ec2022}; Arias et al.\cite{scasp-iclp2018}).

\vspace{-0.15in}

\subsection{FOLD-SE}
\vspace{-0.05 in}
Wang \& Gupta\cite{foldse} devised an efficient, explainable, \textit{rule-based machine learning (RBML)} algorithm for classification tasks known as FOLD-SE. It comes from the FOLD (First Order Learner of Defaults) family of algorithms  (Shakerin et al.\cite{fold}; Wang \& Gupta\cite{foldrpp}). For given input data (numerical and categorical), FOLD-SE generates a set of default rules—essentially a stratified normal logic program—as an (explainable) trained model. The explainability obtained through FOLD-SE is scalable. Regardless of the size of the dataset, the number of learned rules and learned literals stay relatively small while retaining good accuracy in classification when compared to other approaches such as  decision trees. The rules serve as the model, and their accuracy is comparable with traditional tools such as Multi-Layer Perceptrons (MLP) with the added advantage of being explainable. 

\vspace{-0.15 in}
\subsection{The Planning Problem}
\vspace{-0.05 in}
The planning problem involves reasoning about state transitions, goal achievement, and adherence to constraints. It describes finding a sequence of transitions that leads from an initial state to a goal state. In ASP, the planning problem is encoded into a logic program where rules define the transitions and the constraints restrict the allowed transitions. (Gelfond \& Kahl\cite{gelfond-kahl}). In ASP, solutions are represented as answer sets (possible worlds that satisfy constraints). A path/plan describes a series of transitions from the initial state/world to the goal state/world through intermediate possible worlds.
Possible worlds/states are often represented as a set of facts or logical predicates. Solving the planning problem involves searching through the space of all possible transitions to find a path that satisfies the conditions of the goal state while adhering to constraints.
\vspace{-0.2in}
\section{Motivation}
\vspace{-0.05 in}
Consider the situation where an individual was assigned an undesired decision. For example, when applying for a loan, the loan application is denied. The individual has two choices: 1) accept the undesired decision (loan rejection) or 2) see what changes he/she can make to obtain a desired decision/ outcome. Our approach \textit{CFGs} guides individuals by informing them of the changes required to flip the undesired decision to a desired one.


\begin{table*}[t]
\centering
\fontsize{8}{10}\selectfont
\setlength{\tabcolsep}{2.5pt}

\begin{tabular}{@{}p{2.8cm} p{2.3cm} p{1cm} p{2.3cm} p{1cm} p{2.3cm}@{}} 
\toprule
Features & Initial\_State & Action & Intermediate\_State & Action & Goal\_State\\ 
\midrule
Checking account status &no checking account& N/A &no checking account& \textbf{Direct}& $>1000$\\ 
\cmidrule(lr){1-6}
Credit history&all dues at bank cleared& N/A &all dues at bank cleared& N/A &all dues at bank cleared\\ 
\cmidrule(lr){1-6}
Property & no property & N/A & no property & N/A  & no property \\ 
\cmidrule(lr){1-6}
Duration months & 7 & \textbf{Direct} & $>7$ and $\leq 72$ & N/A & $>7$ and $\leq 72$\\ 
\cmidrule(lr){1-6}
Credit amount & 300 & N/A & 300 & N/A & 300\\ 

\bottomrule
\end{tabular}
\caption{Table showing a Path of length 3 showing 2 transition to reach the counterfactual state. Here a direct action is done on feature \textit{Duration month} to increase its value from 7 to the range $>7$ and $\leq 72$. Additionally, for the intermediate state, a direct action is done on feature \textit{Checking account status} to change its value to $>1000$ }
\label{tbl_example_path}
\end{table*}


In the process of flipping an undesired decision to a desired decision, our \textit{CFGs} approach will model--- 1) a solution/world that represents an undesired outcome, e.g., a loan is denied(\textit{initial state $I$}), and 2) a collection of alternative imaginary scenario(s)/world(s) where, after making interventions, we obtain the desired outcome, e.g., the loan is approved, (\textit{a set of goal states $G$}). The initial state $I$ and goal set $G$ are characterized by the decisions made by the model (e.g., the loan approval system). 

Recall that, in the loan approval example; we are denied a loan in the \textit{initial state $I$}, while in every scenario in the \textit{goal set $G$}, we obtain the loan. So, the  decision in the \textit{initial state $I$} should not hold for any scenario in the \textit{goal set $G$}. 
Thus, the original query {\tt ?- approve\_loan(john)} is {\tt False} in the \textit{initial state $I$} and {\tt True} for the \textit{goal set $G$}. 
Through our approach \textit{CFGs}, we aim to symbolically compute the interventions/transitions needed to flip a decision. The interventions/transitions can be thought of as finding a path to a counterfactual state that results in a flipped decision. 
We use the s(CASP) query-driven predicate ASP system (Arias et al.\cite{scasp-iclp2018}) to achieve this goal.
The advantage of s(CASP) lies in automatically generating dual rules, i.e., rules that let us constructively execute negated queries (e.g., {\tt not approve\_loan/1} above).

\vspace{-0.15 in}
\subsection{Motivating Example}
\vspace{-0.05 in}
Assume we have a model that gives a label of loan \textit{approved} or loan \textit{rejected}. Suppose John applies for a loan and is rejected; our goal is to make changes to feature values in John's record so that the loan will be approved. Table \ref{tbl_example_path} shows a path describing the changes that John must make to have his loan approved. The first intervention/action suggests altering the feature \textit{Duration months} from a value of \textit{7} to a value in the range $>7\ and \leq72$. Unfortunately, John is still unsuccessful in getting the loan approved despite making the change. Now \textit{CFGs} suggests intervening on the feature \textit{Checking account status} by changing the feature value \textit{no checking account} to having a checking account with a balance $>1000$, essentially informing John to open a bank account and deposit $1000$ Deutsche marks in it. By making these two changes, the model now predicts that John's loan should be approved.

\begin{figure}[htp]
    \centering
    \includegraphics[width=13cm]{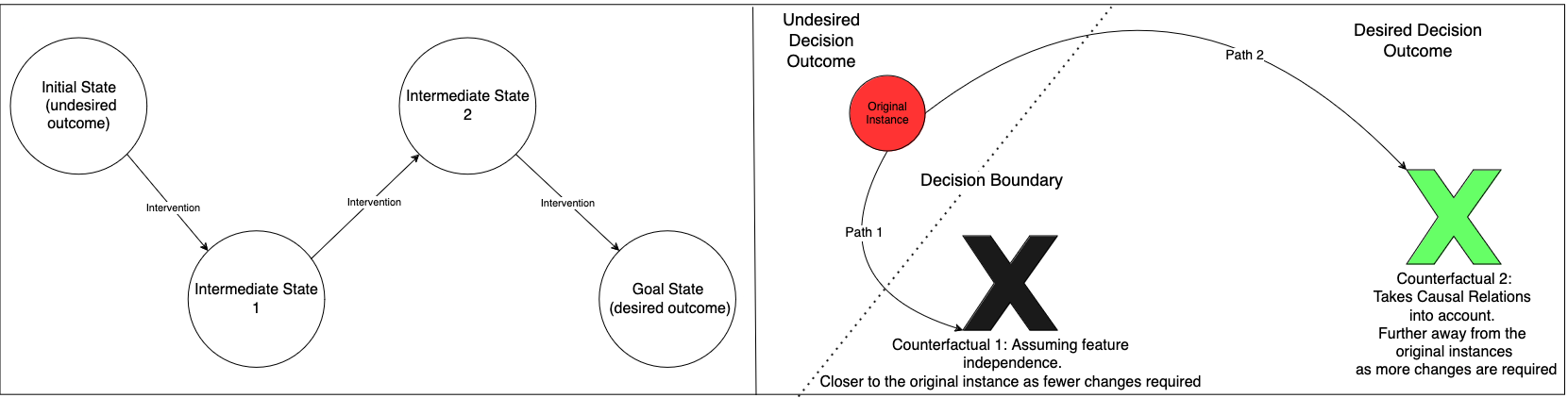}
    \caption{\textbf{Left:} Path showing a series of interventions to go from the original outcome to the desired outcome. \textbf{Right:} Interventions take the original instance to the other side of the decision boundary. Considering causal dependencies, the new counterfactual is further away as more changes are made to the original instance.}
    \label{fig_1}
\end{figure}

\vspace{-0.45 in}
\section{Methodology}
\vspace{-0.05 in}
One of the advantages of \textit{CFGs} is that for the original decision-making rules that result in an undesired outcome, 1) causal rules are used to generate causally consistent counterfactual instances (realistic instances), and 2) a path is returned that shows the actions taken to reach the realistic counterfactual instance.

In this section, we specify the methodology employed by \textit{CFGs} for generating a path to the counterfactual instances/goal states (that produce a \textit{desired outcome}) from an original instance/initial state (that produces an \textit{undesired outcome}). In order to explain our methodology, we must first define specific terms.

\vspace{-0.15in}
\subsection{Theoretical Definitions}
\vspace{-0.1 in}
\begin{definition}
State Space: Let \( D_1, ..., D_n \) represent the domains of the features \( 1, ..., n\). Then the state space is expressed as \( S \) is a set of possible states where each state is defined as a tuple of feature values \( V_1, ..., V_n \)
\begin{equation}
S = \{ (V_1,V_2,...,V_n) | V_i \in D_i,\ for\ each\ i\ in\ 1,...,n \} \label{defn:1}
\end{equation}
\end{definition}
\vspace{-0.15in}

\begin{definition}\label{defn:S_C}
Causally Consistent State Space:
$C$ represents a set of causal rules among the features within a state space $S$. Then,
$\theta_C: P(S) \rightarrow P(S)$ (where $P(S)$ is the power set of S) is a function that defines the subset of a given state sub-space $S'\subseteq S$.
\begin{equation}
\theta_C(S') = \{ s \in S' \mid s\ satisfies\ all\ causal\ rules\ in\ C\} \label{defn:2_1}
\end{equation}
\vspace{-0.05 in}
Given $S$ and $\theta_C$, we define the causally consistent state space $S_C$ as 
\begin{equation}
S_C = \theta_C(S) \label{defn:2_2}
\end{equation}

\end{definition}
\vspace{-0.15 in}

\begin{definition}\label{defn:S_Q}
Decision Consistent State Space:
$Q$ represents a set of rules that compute some external decision for a given state.
$C$ represents a set of causal rules among the features within a state space $S$.
$\theta_Q: P(S) \rightarrow P(S)$ is a function that defines the subset of the causally consistent state space  $S'\subseteq S_C$  that is also consistent with the set of decision rules in~$Q$
\begin{equation}
\theta_Q(S') = \{ s \in S' \mid s\ satisfies\ any\ decision\ rule\ in\ Q\} \label{defn:3_1}
\end{equation}
Given $S_C$ and $\theta_Q$, we define the decision consistent state space $S_Q$ as 
\begin{equation}
S_Q = \theta_Q(S_C) = \theta_Q(\theta_C(S)) \label{defn:3_2}
\end{equation}
\end{definition}

\vspace{-0.15 in}
\begin{definition}\label{defn:I}
Initial State:
The initial state $I$ is an element of the \textit{causally consistent state space} $S_C$
\vspace{-0.05 in}
\begin{equation}
I \in S_C \label{defn:4}
\end{equation}
\end{definition}

\vspace{-0.17 in}

\begin{definition}\label{defn:delta}
Transition Function:
A transition function $\delta: S_C\rightarrow P(S_C)$ maps a causally consistent state to the set of allowable causally consistent states that can be transitioned to in a single step.


\end{definition}

\vspace{-0.1 in}

\begin{definition}[\textbf{Counterfactual Generation Problem (CFG)}]\label{problem_statement}
A counterfactual generation problem (CFG): For the causally consistent state space $S_C$, the decision consistent state space $S_Q$, the initial state $I\in S_C$, and the transition function $\delta$, collectively represented as $(S_C,S_Q,I,\delta)$.
\end{definition}
\vspace{-0.1 in}

\begin{definition}\label{defn:G}
Goal Set: Given a CGF given by $(S_C,S_Q,I,\delta)$, the goal set $G$ is a \textit{causally consistent state space}, $G \subseteq  S_C$, such that elements of $G$ do \textbf{not} satisfy the set of decision rules $Q$.
\vspace{-0.03 in}
\begin{equation}
G = \{ s \in S_C| s\not\in S_Q\} \label{defn:5}
\end{equation}
\end{definition}
\vspace{-0.1 in}

\begin{definition}[\textbf{Solution Path}]\label{solution_path}
Solution Path: A solution to the problem $(S_C,S_Q,I,\delta)$ with Goal set $G$ is a path:
\vspace{-0.05 in}
\begin{equation} \label{defn:8}
\begin{split}
s_0,s_1,...s_m\ where\ s_j\in S_C\ for\ all\ j\ \in\{0,...,m\}\ such\ that \\
s_0,...,s_{m-1} \not\in G;\ s_{i+1}\in \delta(s_i)\ for\ i\ \in\{0,...,m-1\};\ s_0 = I; s_m\in G 
\end{split}
\end{equation}
\end{definition}

\vspace{-0.2 in}

\subsection{Methodology Definitions}

\vspace{-0.1 in}

\begin{definition}[Rule]\label{defn:rule}
    A predicate that takes as input a state $s$ and give us $TRUE$/$FALSE$ as an output. A state $s$ satisfies a rule $r$ if and only if $r(s)$ returns $TRUE$.
\end{definition}

\vspace{-0.05 in}

\begin{definition}[Action]\label{defn:action}
    A function that takes as input a state $s$ and give us as output a state $s'$ where $s\not=s'$
\end{definition}

\vspace{-0.1 in}
\begin{definition}[\textbf{not\_member}]
Function \textbf{not\_member}: checks if a specific element is both: 1) \textbf{not} a member of a list, and 2) given a list of tuples, \textbf{not} a member of any tuple in the list
\end{definition}



\vspace{-0.1 in}
\begin{definition}[\textbf{get\_last}]
Function \textbf{get\_last}: Returns the last member of a list.
\end{definition}

\vspace{-0.1 in}
\begin{definition}[\textbf{pop}]
Function \textbf{pop}: Removes and returns the last member of a list.
\end{definition}

\vspace{-0.25 in}

\subsection{Algorithm}
\vspace{-0.05 in}

\subsubsection{Obtain a path to the counterfactual state}
\begin{algorithm}[h!]
\caption{\textbf{get\_path}: Obtain a path to the counterfactual state}
\label{alg_path}
\begin{algorithmic}[1]
    \REQUIRE Initial State $(I)$, States $S$, Causal Rules $C$, Decision Rules $Q$, \textit{is\_counterfactual} (Algorithm \ref{alg_counterfactual}), \textit{delta} (Algorithm \ref{alg_intervene}), Actions $a\in A$:
    \begin{itemize}
    \item Causal Action: $s$ gets altered to a causally
    consistent new state $s'=a(s)$. OR 
    \item Direct Action: new state $s'=a(s)$ is obtained by altering 1 feature value of $s$.     
    \end{itemize}
    \STATE\label{PS21} Create an empty list \textit{visited\_states} that tracks the list of states traversed (so that we avoid revisiting them).
    \STATE Append ($I$, [~]) to \textit{visited\_states} 
    \WHILE{$is\_counterfactual(get\_last(visited\_states),C,Q)\ is\ FALSE$ }
    \STATE Set $visited\_states=intervene(visited\_states,C,A) $
    \ENDWHILE
    \STATE \textit{visited\_states}

\end{algorithmic}
\end{algorithm}
\vspace{-0.03 in}
The function \textit{get\_path} is our algorithmic implementation of obtaining the Solution Path $P$ from definition \ref{solution_path}. In Algorithm \ref{alg_path}, we specify the pseudo-code for a function \textit{get\_path}, which takes as arguments an Initial State $I$ that is causally consistent, a set of Causal Rules $C$, a set of decision rules $Q$, and a set of actions $A$. 
The function returns a path to the counterfactual state/goal state $g\in G$ for a given Initial State $I$ (original individual) in the form of a list \textit{candidate\_path}.

Initially, the current state $s=I$. Given the current state $I$, the function \textbf{get\_path} checks if $s$ is a counterfactual in line 3. If the initial state $I$ is already a counterfactual, \textbf{get\_path} returns a path in the form of a list containing $I$. If $I$ is not a counterfactual, the algorithm traverses from $s=I$ to a new causally consistent state $s'$ using a transition function \textbf{intervene} and updates the list \textit{visited\_states} with the new state $s'$. The new state $s'$, by definition, is causally consistent. The algorithm checks again if the new state $s'$ is a counterfactual by seeing if the function \textbf{\textit{is\_counterfactual}} satisfies $s'$ (line 3). If yes, the algorithm returns the list \textit{visited\_states} which contains the path from $I$ to $s'$. If not, it updates the $current\_state$ to the new state $s'$ and keeps repeating these steps until it reaches a counterfactual/goal state $g$, i.e., it keeps running until $s'=g$. The algorithm ends when a counterfactual state is reached, i.e., \textbf{\textit{is\_counterfactual}} satisfies $s'$,

\vspace{-0.15 in}
\begin{equation}
\begin{split}
   s'\in G \mid\ by\ definition \\
   s_0,...,s_k,s' \mid s_0,...,s_k: \not\in G
\end{split}
\end{equation}





\vspace{-0.22 in}
\subsubsection{Checking for Goal State/ Counterfactual State:}
\vspace{-0.1 in}
\begin{algorithm}[h!]
\caption{\textbf{is\_counterfactual}: checks if a state is a counterfactual/goal state}
\label{alg_counterfactual}
\begin{algorithmic}[1]
    \REQUIRE State $s\in S$, Set of Causal \textit{rules} $C$, Set of Decision \textit{rules} $Q$
    \IF{$s$ satisfies \textbf{ALL} rules in $C$ \textbf{AND} $s$ satisfies \textbf{NO} rules in $Q$}
    \STATE Return $TRUE$.
    \ELSE    
    \STATE Return $FALSE$.
    \ENDIF
\end{algorithmic}
\end{algorithm}
\vspace{-0.05 in}
The function \textit{get\_counterfactual} is our algorithmic implementation of checking if a state $s\in G$ from definition \ref{defn:G}.
In Algorithm \ref{alg_counterfactual}, we specify the pseudo-code for a function \textit{is\_counterfactual} which takes as arguments a state $s\in S$, a set of causal rules $C$, and a set of Decision rules $Q$. The function checks if a state $s\in S$ is a counterfactual/goal state. By definition \textit{is\_counterfactual} is $TRUE$ for state $s$ that is causally consistent with all $c\in C$ and \textbf{does not} agree with the any decision rules $q\in Q$.

\vspace{-0.15 in}
\begin{equation}
   is\_counterfactual(s,C,Q)=TRUE \mid s\ agrees\ with\ C;\ s\ disagrees\ with\ Q; 
\end{equation}

\vspace{-0.25 in}
\subsubsection{Transition Function: Moving from the current state to the new state}
\vspace{-0.1 in}
\begin{algorithm}[h!]
\caption{\textbf{intervene}: reach a causally consistent state from a causally consistent current state}
\label{alg_intervene}
\begin{algorithmic}[1]
    \REQUIRE Causal \textit{rules} $C$, List \textit{visited\_states}, List \textit{actions\_taken}, Actions $a\in A$:
    \vspace{-0.1 in}
    \begin{itemize}
    
    \item Causal Action: $s$ gets altered to a causally
    consistent new state $s'=a(s)$. OR 
    \item Direct Action: new state $s'=a(s)$ is obtained by altering 1 feature value of $s$.     
    \end{itemize}
    \vspace{-0.1 in}

    \STATE Set $(s, actions\_taken)$ = \textit{pop(visited\_states)}
    \STATE Try to select an action $a\in A$ ensuring \textit{not\_member(a(s),visited\_states)} and \textit{not\_member(a,actions\_taken)} are $TRUE$
    \IF{ $a$ exists}
    \STATE Set $(s, actions\_taken),visited\_states=update(s,visited\_states,actions\_taken,a)$
    \ELSE
    \STATE //Backtracking 
    \IF {\textit{visited\_states} is empty }
    \STATE \textit{EXIT with Failure}
    \ENDIF
    \STATE Set $(s, actions\_taken)$ = \textit{pop(visited\_states)}
    \ENDIF
    \STATE Set $(s,actions\_taken), visited\_states=$\\ \hspace{0.2 in} $make\_consistent(s,actions\_taken, visited\_states,C,A)$ 
    \STATE Append $(s,actions\_taken)$ to \textit{visited\_states}
    \STATE Return \textit{visited\_states}.

\end{algorithmic}
\end{algorithm}
\vspace{-0.1 in}
The function \textit{intervene} is our algorithmic implementation of the transition function $\delta$ from definition \ref{defn:delta}. In Algorithm \ref{alg_intervene}, we specify the pseudo-code for a function \textit{intervene}, which takes as arguments an Initial State $I$ that is causally consistent, a set of Causal Rules $C$, and a set of actions $A$. It is called by \textit{get\_path} in line 4 of algorithm \ref{alg_path}.

The function \textit{intervene} acts as a transition function that inputs a list \textit{visited\_states} containing the current state $s$ as the last element, and returns the new state $s'$ by appending $s'$ to \textit{visited\_states}. The new state $s'$ is what the current state $s$ traverses. Additionally, the function \textit{intervene} ensures that no states are revisited. In traversing from $s$ to $s'$, if there are a series of intermediate states that are \textbf{not} causally consistent, it is also included in \textit{visited\_states}, thereby depicting how to traverse from 1 causally consistent state to another.


\vspace{-0.2 in}

\subsubsection{Make Consistent}
\begin{algorithm}[h!]
\caption{\textbf{make\_consistent}: reaches a consistent state}
\label{alg_inner_delta}
\begin{algorithmic}[1]
    \REQUIRE State $s$, Causal \textit{rules} $C$, List \textit{visited\_states} , \textit{actions\_taken}, Actions $a\in A$:
    \begin{itemize}
    \vspace{-0.1 in}
    \item Causal Action: $s$ gets altered to a causally
    consistent new state $s'=a(s)$. OR 
    \item Direct Action: new state $s'=a(s)$ is obtained by altering 1 feature value of $s$.     
    \end{itemize}
    \vspace{-0.1 in}
    \WHILE{$s$ does not satisfy all rules in $C$}
    \STATE Try to select a causal action $a$ ensuring \textit{not\_member(a(s),visited\_states)} and \textit{not\_member(a,actions\_taken)} are $TRUE$
    \IF{ causal action $a$ exists}
    \STATE Set $(s, actions\_taken),visited\_states=update(s,visited\_states,actions\_taken,a)$
    \ELSE
        \STATE Try to select a direct action $a$ ensuring \textit{not\_member(a(s),visited\_states)} and \textit{not\_member(a,actions\_taken)} are $TRUE$
        \IF{ direct action $a$ exists}
        \STATE Set $(s, actions\_taken), visited\_states=update(s,visited\_states,actions\_taken,a)$
        \ELSE
        \STATE //Backtracking 
        \IF {\textit{visited\_states} is empty }
        \STATE \textit{EXIT with Failure}
        \ENDIF
        \STATE Set $(s, actions\_taken)$ = \textit{pop(visited\_states)}
        \ENDIF
    \ENDIF
    \ENDWHILE
    \STATE Return $(s,actions\_taken),visited\_states$ .

\end{algorithmic}
\end{algorithm}
In Algorithm \ref{alg_inner_delta}, we specify the pseudo-code for a function \textit{make\_consistent}. It takes as arguments a current State $s$, a list \textit{actions\_taken}, a list \textit{visited\_states}, a set of Causal Rules $C$ and a set of actions $A$. It is called by the function \textit{intervene} in line 12 of algorithm \ref{alg_intervene}.

The function \textit{make\_consistent} is a  sub-function of \textit{intervene} that does the action task of transitioning from the current state to a new state that is causally consistent.

\vspace{-0.15 in}
\subsubsection{Update}
\begin{algorithm}[h!]
\caption{\textbf{update}: Updates the list \textit{actions\_taken} with the planned action. Then updates the current state.}
\label{alg_update}
\begin{algorithmic}[1]
    \REQUIRE State $s$, List \textit{visited\_states}, List \textit{actions\_taken}, Action $a \in A$:
    \vspace{-0.1 in}
    \begin{itemize}
        \item Causal Action: $s$ gets altered to a causally
        consistent new state $s'=a(s)$. OR 
        \item Direct Action: new state $s'=a(s)$ is obtained by altering 1 feature value of $s$.  
    \end{itemize}
    \vspace{-0.1 in}  
    \STATE Append $a$ to \textit{actions\_taken}.
    \STATE Append $(s, actions\_taken)$ to \textit{visited\_states}.
    \STATE Set $s=a(s)$.
    \RETURN $(s, [~])$, \textit{visited\_states}

\end{algorithmic}
\end{algorithm}
\vspace{-0.2 in}
In Algorithm \ref{alg_update}, we specify the pseudo-code for a function \textit{update}, that given a state $s$, list \textit{actions\_taken}, list \textit{visited\_states}and given an action $a$, appends $a$ to \textit{actions\_taken}. It also appends the list  \textit{actions\_taken} as well as the new resultant state resulting from the action $a(s)$ to the list  \textit{visited\_states}. The list \textit{actions\_taken} is used to track all the actions attempted from the current state to avoid repeating them. The function \textit{update} is called by both functions \textit{intervene} and \textit{make\_consistent}.

\vspace{-0.15 in}
\subsection{Plausibility Constraints}
\vspace{-0.05 in}

There are cases where certain feature values may be immutable or restricted with respect to changes being made, e.g., \textit{age} cannot decrease or \textit{credit score} cannot be directly altered. To ensure that such restrictions to these features are respected, we introduce \textit{plausibility constraints}. We can think of these plausibility constraints being applied to the actions in our algorithms, as it is through the actions taken that changes are made to the features. Additionally since these \textit{plausibility constraints} \textit{do not} add any new states but restrict the states that can be reached from the resulting actions, they are represented in the algorithms \ref{alg_path},  \ref{alg_intervene},  \ref{alg_inner_delta},  \ref{alg_update} through the set of available actions.
\vspace{-0.15in}

\subsection{Soundness}
\vspace{-0.05 in}
\begin{definition}[CFG Generation]\label{defn:impl_cfg}
We define our implementation of the CFG as defined in definition \ref{problem_statement}.
Suppose we run algorithm \ref{alg_path} with inputs:
        Initial State $I$ (Definition \ref{defn:2_2}),        
        States Space $S$ (Definition \ref{defn:2_1}),        
        Set of Causal Rules $C$ (Definition \ref{defn:rule}),        
        Set of Decision Rules $Q$ (Definition \ref{defn:rule}), and        
        Set of Actions $A$ (Definition \ref{defn:action}).

\textbf{Direct Action vs. Causal Action:}
Every Direct action modifies a single field of a state and every causal action results in a causally consistent state with respect to $C$.
    
We can define a CFG with causally consistent state space $S_C$ (Definition \ref{defn:S_C}), Decision consistent state space $S_Q$ (Definition \ref{defn:S_Q}), Initial State $I$, transition function:

\vspace{-0.1 in}
\[\{(s,\{s' | s' = \sigma(s,a), a\in A, s'\in S_C \}) | s \in S_C\},\textrm{ where}\] 
\vspace{-0.2 in}
  \[\sigma(s,a) = \left\{\begin{array}{l}
     a(s) \textit{ if } a(s) \in S_C\\
     \sigma(a(s),a') \textit{ with } a\in A\textit{, otherwise}
  \end{array}\right.\]

\end{definition}

\vspace{-0.1 in}
\begin{definition}\label{defn:candidate_path}

\textit{Candidate path}: Given the CFG $(S_C,S_Q,I,\delta)$ constructed from a run of  algorithm \ref{alg_path} and the return value that we refer to as $r$ such that r is a list (algorithm succeeds). $r'$ is the resultant list obtained from removing all elements containing states $s'\not\in S_C$. We can construct the corresponding candidate path as follows: $r'_i$ represents the $i$ th element of list $r'$. The candidate path is the sequence of states $s_0,...,s_{m-1}$, where $m$ is the length of list $r'$. $s_i$ is the state corresponding to $r'_i$ for $0\leq i<m$.       
\end{definition}

Definition \ref{defn:impl_cfg} defines how to map the input of algorithm \ref{alg_path} to a CGF (definition \ref{problem_statement}). Definition \ref{defn:candidate_path} defines how to map the result of algorithm \ref{alg_path} to a possible solution of the corresponding CFG. From theorem \ref{theorem_soundness} \textit{(proof in supplement)}, the \textit{candidate path} (definition \ref{defn:candidate_path}) is a solution to the corresponding \textit{CFG problem} (definition \ref{defn:impl_cfg}).\\

\vspace{-0.1 in}
\noindent \textit{Theorem 1}\\ 
Soundness Theorem\\
Given a CFG $\mathbb{X}=(S_C,S_Q,I,\delta)$, constructed from a run of algorithm \ref{alg_path} and a corresponding candidate path $P$, $P$ is a solution path for $\mathbb{X}$.

\vspace{-0.2 in}
\section{Experiments}\label{Experiments}
\vspace{-0.05 in}
We have applied our \textit{CFGs} methodology to rules generated by \textit{RBML} algorithm FOLD-SE. For our experiments we used the Adult dataset(\cite{adult}),  the Car Evaluation dataset(\cite{car}), and the German dataset (\cite{german}). 
The Adult dataset contains individuals' demographic information with the label indicating whether someone makes `$=<$\$50k/year' or `$>$\$50k/year'. The German dataset contains demographic information of individuals with the label indicating whether someone's credit risk is \textit{`good'} or \textit{`bad'}. The Car Evaluation dataset \cite{car} contains information on the acceptability of purchasing a car. 

\vspace{-0.15in}
\subsection{Obtaining Paths}
\vspace{-0.05 in}

In our experiments on the adult dataset, our \textit{RBML} algorithm gives us rules indicating whether someone makes `$=<$\$50k/year'. Given that the decision-making rules obtained from these datasets specify an undesired outcome (a person making less than or equal to \$50K) for an original instance, the goal is to find a path to a counterfactual instance where a person makes more than \$50K. Like the Adult dataset, rules are obtained if someone has a good or bad credit from the German dataset where the undesired outcome is the label corresponding to \textit{good} credit rating. For the Car Evaluation dataset, the decision rules inform us if the car is acceptable to buy or reject where the undesired outcome is the label corresponding to \textit{reject}.

Our \textit{CFGs} methodology produces a list of \textit{original-counterfactual} pairs that denote all the possible paths to be taken by an original instance in reaching a counterfactual. We have shown our results for the German dataset in Table \ref{tbl_german} for a path from someone who has a 'good' credit risk rating to having a 'bad' rating. For the imbalanced German dataset, the rules obtained identified individuals with a good rating, and hence we found the counterfactuals of those individuals, i.e., individuals with a bad rating and the path depicts the steps to avoid for a person to go from having a \textit{`good'} credit risk rating to having a \textit{`bad'} credit risk rating. The Adult dataset in Table \ref{tbl_adult} shows a path from the original instance that makes `$=<$\$50k/year' to a counterfactual state that makes `$>$\$50k/year'. The car evaluation dataset in Table \ref{tbl_cars} shows a path for a car that would be \textit{rejected} initially, but the changes suggested make the car \textit{acceptable} for purchase.

\begin{table*}[t]
\centering
\fontsize{6}{8}\selectfont
\setlength{\tabcolsep}{2.5pt}

\begin{tabular}{@{}p{2 cm} p{2 cm} p{1cm} p{3 cm}@{}} 
\toprule
Features & Initial\_State & Action & Goal\_State\\ 
\midrule
Marital\_Status & never\_married & N/A & never\_married \\ 
\cmidrule(lr){1-4}
Capital Gain & \$1000 & \textbf{Direct} &$> 6849$ and $\leq 99999$\\ 
\cmidrule(lr){1-4}
Education\_num & $11$ & N/A & $11$\\ 
\cmidrule(lr){1-4}
Relationship & unmarried & N/A & unmarried\\ 
\cmidrule(lr){1-4}
Sex & male & N/A & male\\ 
\cmidrule(lr){1-4}
Age & 28 & N/A & 28\\ 

\bottomrule
\end{tabular}
\caption{\textit{adult}- Path of length 2 shows one transition to reach the goal state. The feature value of \textit{Capital Gain} goes from $\$1000$ to the range $> \$6849$ and $\leq \$99999$.}
\label{tbl_adult}
\end{table*}


\begin{table*}[t]
\centering
\fontsize{6}{8}\selectfont
\setlength{\tabcolsep}{2.5pt}

\begin{tabular}{@{}p{1.5cm} p{2 cm} p{1cm} p{2 cm}@{}} 
\toprule
Features & Initial\_State & Action & Goal\_State\\ 
\midrule
persons & 2 & \textbf{Direct} & 4 \\ 
\cmidrule(lr){1-4}
maint & medium& N/A &medium\\ 
\cmidrule(lr){1-4}
buying & medium & N/A & medium\\ 
\cmidrule(lr){1-4}
safety & medium & N/A & medium\\ 

\bottomrule
\end{tabular}
\caption{\textit{car evaluation}- Path of length 2 shows one transition to reach the goal state. The feature value of \textit{persons} goes from $2$ to the $4$.}
\label{tbl_cars}
\end{table*}

\begin{table*}[t]
\centering
\fontsize{6}{8}\selectfont
\setlength{\tabcolsep}{2.5pt}

\begin{tabular}{@{}p{2.5cm} p{2.3cm} p{1cm} p{2.3cm} p{1cm} p{2.3cm}@{}} 
\toprule
Features & Initial\_State & Action & Intermediate\_State & Action & Goal\_State\\ 
\midrule
Checking account status &no checking account& N/A &no checking account& \textbf{Direct} & $\geq200$\\ 
\cmidrule(lr){1-6}
Credit history&all dues at bank cleared& N/A &all dues at bank cleared& N/A &all dues at bank cleared\\ 
\cmidrule(lr){1-6}
Property & car or other & N/A & 11 & N/A  & 11 \\ 
\cmidrule(lr){1-6}
Duration months & 7 & \textbf{Direct} & $>7$ and $\leq 72$ & N/A & $>7$ and $\leq 72$\\ 
\cmidrule(lr){1-6}
Credit amount & 300 & N/A & 300 & N/A & 300\\ 
\cmidrule(lr){1-6}
Job & official/skilled employee & N/A & official/skilled employee & N/A & official/skilled employee\\ 
\cmidrule(lr){1-6}
Present Employment Since & $\geq 1$ and $<4$ & N/A & $\geq 1$ and $<4$& N/A & $\geq 1$ and $<4$\\ 

\bottomrule
\end{tabular}
\caption{\textit{german}- Path of length 3 shows two transitions to reach the goal state. The value of \textbf{Duration month} increases from $7$ to the range $>7$ and $\leq 72$. Then, the value of \textbf{Checking account status} of the intermediate state changes to $>200$}
\label{tbl_german}
\end{table*}

\vspace{-0.1 in}

\vspace{-0.15in}
\section{Related Work}
\vspace{-0.05 in}

There are existing approaches to tackle the problem of transparency by utilizing counterfactuals, such as that of Wachter et al.\cite{wachter}. Some approaches are tied to particular models or families of models, while others are optimization-based approaches such as Tolomei et al.\cite{ref_mace_1}) and Russel\cite{ref_mace_2}. Ustun et al.\cite{ref_2_ustun} took an approach that highlighted the need to focus on algorithmic recourse  to ensure a viable counterfactual explanation. White \& Garcez\cite{ref_clear} demonstrated how to utilize counterfactual explanations to improve model performance and ensure accurate explanations. Lately, specific approaches have considered type of features being altered and focused on producing viable realistic counterfactuals, such as the work by Karimi et al.\cite{alt_karimi}, which has the additional advantage of being model-agnostic. The work done by Karimi et al.\cite{ref_4_karimi_2} showed how generation of counterfactuals should consider causal rules to make them realizable. They introduce the idea of interventions as a solution to achieving the counterfactual state.The approaches of Bertossi \& Reyes\cite{ref_asp_cf} utilize ASP, however, it does not use a goal-directed ASP system and relies on grounding, which has the disadvantage of losing the association between variables. On the other hand \textit{Counterfactual Generation with s(CASP) (CFGs)} utilizes s(CASP); hence, no grounding is required. It justifies decisions through counterfactual explanations by utilizing the answer set programming paradigm (ASP) while providing a step-by-step path of going from an undesired outcome to a desired outcome. It has the flexibility to generate counterfactual instances regardless of the \textit{RBML} algorithm. In the future, we plan on incorporating counterfactual reasoning to help with explainability in image classification, similar to the work done by Parth et al.\cite{parth_nesy} and Parth et al.\cite{parth_padl}.


\vspace{-0.2in}
\section{Conclusion and Future  Work}

The main contribution of this paper is the \textit{Counterfactual Generation with s(CASP) (CFGs)} framework that shows that complex tasks, such as imagining possible scenarios, which is essential in generating counterfactuals, as shown by Byrne\cite{ref_Byrne_CF} (``What if something else happened?"), can be modeled with s(CASP) while considering causal dependencies between features. Such imaginary worlds/states (where alternate facts are true and different decisions are made) can be automatically computed using the s(CASP) system. It also addresses the problem of generating a path to obtaining counterfactual explanations. \textit{CFGs} shows how ASP and specifically the s(CASP) goal-directed ASP system can be used for generating the path to achieving these counterfactual explanations regardless of the \textit{RBML} algorithm used for generating the decision-making rules. 




\vspace{-0.2in}

\bibliographystyle{tlplike}
\bibliography{bibliography}

\section{Supplimentary Material}

\subsection{Proofs}

\begin{theorem}{Soundness Theorem}\label{theorem_soundness}\\
\noindent Given a CFG $\mathbb{X}=(S_C,S_Q,I,\delta)$, constructed from a run of algorithm \ref{alg_path} and a corresponding candidate path $P$, $P$ is a solution path for $\mathbb{X}$.
\begin{proof}
Let $G$ be a goal set for $\mathbb{X}$. By definition \ref{defn:candidate_path} $P=s_0,...,s_{m}$, where $m\geq0$.
By definition \ref{solution_path}
we must show $P$ has the following properties.
        
        1) $s_0=I $
        
        2) $s_m\in G $
        
        3) $s_j\in S_C\ for\ all\ j\ \in\{0,...,m\}$
        
        4) $s_0,...,s_{m-1} \not\in G$
        
        5) $s_{i+1}\in \delta(s_i)\ for\ i\ \in\{0,...,m-1\}$\\
1) By definition \ref{defn:I}, $I$ is causally consistent and cannot be removed from the candidate path. Hence I must be in the candidate path and is the first state as per line 2 in algorithm \ref{alg_path}. Therefore $s_0$ must be $I$.\\
2) The while loop in algorithm 5 ends if and only if $is\_counterfactual(s,C,Q)$ is True. From theorem 1 $is\_counterfactual(s,C,Q)$ is True only for the goal state. Hence $s_m\in G$.\\
3) By  definition\ref{defn:candidate_path} of the candidate path, all states $s_j\in S_C\ for\ all\ j\ \in\{0,...,m\}$.\\
4) By theorem \ref{theorem_all_but_last}, we have proved the claim $s_0,...,s_{m-1} \not\in G$.\\
5) By theorem \ref{theorem_delta}, we have proved the claim $s_{i+1}\in \delta(s_i)\ for\ i\ \in\{0,...,m-1\}$.\\
Hence we proved the candidate path $P$ (definition \ref{defn:candidate_path}) is a solution path (definition \ref{solution_path}).







\end{proof}
\end{theorem}



\begin{theorem}\label{theorem_is_counterfactual}
\noindent Given a CFG $\mathbb{X}=(S_C,S_Q,I,\delta)$, constructed from a run of algorithm \ref{alg_path}, with goal set $G$, and $s\in S_C$; $is\_counterfactual(s,C,Q)$ will be $TRUE$ if and only if $s\in G$.
\begin{proof}

By the definition of the goal set $G$ we have
\begin{equation}
G = \{ s \in S_C| s\not\in S_Q\} \label{theorem_1_1}
\end{equation}
For $is\_counterfactual$ which takes as input the state $s$, the set of causal rules $C$ and the set of decision rules $Q$ (Algorithm \ref{alg_counterfactual}), we see that by from line 1 in algorithm \ref{alg_counterfactual}, it returns TRUE if it satisfied all rules in $C$ and no rules in $Q$.

By the definition \ref{defn:S_Q}, $s\in S_Q$ \textit{if and only if} it satisfies a rule in $Q$. 
Therefore, $is\_counterfactual(s,C,Q)$ is $TRUE$ if and only if $s\not\in S_Q$ and since $s\in S_C$ and $s\not\in S_Q$ then $s\in G$.

\end{proof}
\end{theorem}

\begin{theorem}\label{theorem_delta}
\noindent Given a CFG $\mathbb{X}=(S_C,S_Q,I,\delta)$, constructed from a run of algorithm \ref{alg_path} and a corresponding candidate path $P=s_0,...,s_{m}$; $s_{i+1}\in \delta(s_i)\ for\ i\ \in\{0,...,m-1\}$

\begin{proof}
This property can be proven by induction on the length of the list \textit{visited\_lists} obtained from Algorithm 5,4,3.\\
\textbf{Base Case}: The list \textit{visited\_lists} from algorithm \ref{alg_path} has length of 1, i.e., [$s_0$]. The property $s_{i+1}\in \delta(s_i)\ for\ i\ \in\{0,...,m-1\}$ is trivially true as there is no $s_{-1}$.\\
\textbf{Inductive Hypotheses}: 
We have a list [$s_0,...,s_{n-1}$] of length $n$ generated from $0$ or more iteration of running the function \textit{intervene} (algorithm \ref{alg_intervene}), and it  satisfies the claim $s_{i+1}\in \delta(s_i)\ for\ i\ \in\{0,...,n-1\}$\\

\textbf{Inductive Step}: If we have a list  [$s_0,...,s_{n-1}$]  of length n and we wish to get element $s_n$ obtained through running another iteration of function \textit{intervene} (algorithm \ref{alg_intervene}). Since [$s_0,...,s_{n-1}$] is  of length n by the inductive hypothesis, it satisfies the property, and it is sufficient to show $s_n\in\delta(s_{n-1})$ where $s_{i+1}\in \delta(s_i)\ for\ i\ \in\{0,...,n-1\}$.\\

The list \textit{visited\_lists} from algorithm \ref{alg_path} has length of $n$. Going from $s_{n-1}$ to $s_n$ involves calling the function \textit{intervene} (algorithm \ref{alg_intervene}) which in turn calls the function \textit{make\_consistent} (algorithm \ref{alg_inner_delta}).

Function \textit{make\_consistent} (algorithm \ref{alg_inner_delta}) takes as input the state $s$, the list of actions taken \textit{actions\_taken}, the list of visited states \textit{visited\_states}, the set of causal rules $C$ and the set of possible actions $A$.
It returns \textit{visited\_states} with the new causally consistent states as the last element. From line 1, if we pass as input a causally consistent state, then function \textit{make\_consistent} does nothing. On the other hand, if we pass a causally inconsistent state, it takes actions to reach a new state. Upon checking if the action taken results in a new state that is causally consistent from the \textit{while} loop in line 1, it returns the new state. 
Hence, we have shown that the moment a causally consistent state is encountered in function \textit{make\_consistent}, it does not add any new state.

Function \textit{intervene} (algorithm \ref{alg_intervene}) takes as input the list of visited states \textit{visited\_states} which contains the current state as the last element, the set of causal rules $C$ and the set of possible actions $A$. It returns \textit{visited\_states} with the new causally consistent states as the last element. It calls the function \textit{make\_consistent}. For the function \textit{intervene}, in line 1 it obtains the current state (in this case $s_{n-1}$) from the list \textit{visited\_states}. It is seen in line 2 that an action $a$ is taken: 

        1) Case 1: If a causal action is taken, then upon entering the the function \textit{make\_consistent} (algorithm \ref{alg_inner_delta}), it will not do anything as causal actions by definition result in causally consistent states.         

        2) Case 2: If a direct action is taken, then the new state that may or may not be causally consistent is appended to \textit{visited\_states}. The call to the function \textit{make\_consistent} will append one or more states with only the final state appended being causally consistent.

Hence we have shown that the moment a causally consistent state is appended in function \textit{intervene}, it does not add any new state. This causally consistent state is $s_n$. In both cases $s_n=\sigma(s_{n-1})$ as defined in definition \ref{defn:impl_cfg} and this $s_n \in \delta(s_{n-1})$.




\end{proof}
\end{theorem}

\begin{theorem}\label{theorem_all_but_last}
\noindent Given a CFG $\mathbb{X}=(S_C,S_Q,I,\delta)$, constructed from a run of algorithm \ref{alg_path}, with goal set $G$ and a corresponding candidate path $P=s_0,...,s_{m}$ with $m\geq0$, $s_0,...,s_{m-1} \not\in G$.
\begin{proof}
This property can be proven by induction on the length of the list \textit{visited\_lists} obtained from Algorithm 5,4,3.\\
\textbf{Base Case}: \textit{visited\_lists} has length of 1. 
Therefore the property $P=s_0,...,s_{m}$ with $m\geq0$, $s_0,...,s_{m-1} \not\in G$ is trivially true as state $s_{j}$ for $j<0$ does not exist.\\


\textbf{Inductive Hypotheses}: 
We have a list [$s_0,...,s_{n-1}$] of length $n$ generated from $0$ or more iteration of running the function \textit{intervene} (algorithm \ref{alg_intervene}), and it  satisfies the claim $s_0,...,s_{n-2} \not\in G$.\\

\textbf{Inductive Step}: Suppose we have a list [$s_0,...,s_{n-1}$] of length n and we wish to append the n+1 th element (state $s_{n}$) by calling the function \textit{intervene}, and we wish to show that that the resultant list satisfies the claim $s_0,...,s_{n-1} \not\in G$. The first n-1 elements ($s_0,...,s_{n-2}$) are not in $G$ as per the inductive hypothesis.

From line 3 in the function \textit{get\_path} (algorithm \ref{alg_path}), we see that to call the function \textit{intervene} another time, the current state (in this case $s_{n-1})$ \textbf{cannot} be a counterfactual, by theorem \ref{theorem_is_counterfactual}. Hence $s_{n-1}\not\in G$ 

Therefore by induction the claim $s_0,...,s_{n-1}\not \in G$ holds.
\end{proof}
\end{theorem}

\end{document}